\documentclass[conference]{IEEEtran}
\IEEEoverridecommandlockouts
\usepackage{graphics} % for pdf, bitmapped graphics files
\usepackage{epsfig} % for postscript graphics files
\usepackage{mathptmx} % assumes new font selection scheme installed
\usepackage{times} % assumes new font selection scheme installed
\usepackage{microtype}
\usepackage{algorithmic}
\usepackage{textcomp}
\usepackage{xcolor}
\usepackage{gensymb}
\usepackage{caption}
\usepackage{hyperref}
\usepackage{mwe}
\usepackage{verbatim}
\usepackage{amsbsy}
\usepackage{siunitx}
\usepackage{tabularx,booktabs}
\usepackage{dblfloatfix}
\usepackage{cite}
\usepackage{bm}
\usepackage{amsmath,amssymb,amsfonts}
\usepackage{subcaption}
\usepackage{float}

\usepackage{algorithm, algorithmic}
\usepackage[autolanguage]{numprint}

\usepackage{spreadtab}
\usepackage{multirow}% http://ctan.org/pkg/multirow
\usepackage{cite}
\usepackage{hyperref}
\usepackage{xcolor}
\usepackage{algorithmic}
\usepackage{textcomp}
\usepackage{xcolor}
\usepackage{gensymb}
\usepackage{caption}
\usepackage{mwe}
\usepackage{amsfonts}
\usepackage{amssymb}
\usepackage{verbatim}
\usepackage{amsbsy}
\usepackage{siunitx}
\usepackage{tabularx, booktabs}
\usepackage{soul}
\usepackage{tikz}
\usetikzlibrary{calc}
\def\BibTeX{{\rm B\kern-.05em{\sc i\kern-.025em b}\kern-.08em
    T\kern-.1667em\lower.7ex\hbox{E}\kern-.125emX}}

\makeatletter
\newif\if@anonymize

\@anonymizetrue    % Uncomment to hide text
\if@anonymize
  \newcommand{\highlight@DoHighlight}{
    \fill [outer sep = -15pt, inner sep = 0pt, color=black]
          ($(begin highlight)+(0,8pt)$) rectangle ($(end highlight)+(0,-3pt)$) ;
  }

  \newcommand{\highlight@BeginHighlight}{
    \coordinate (begin highlight) at (0,0) ;
  }

  \newcommand{\highlight@EndHighlight}{
    \coordinate (end highlight) at (0,0) ;
  }

  \newdimen\highlight@previous
  \newdimen\highlight@current
  \newlength{\item@width}

  \DeclareRobustCommand*\anonymize{%
    \SOUL@setup
    \def\SOUL@preamble{%
      \begin{tikzpicture}[overlay, remember picture]
        \highlight@BeginHighlight
        \highlight@EndHighlight
      \end{tikzpicture}%
    }%
    \def\SOUL@postamble{%
      \begin{tikzpicture}[overlay, remember picture]
        \highlight@EndHighlight
        \highlight@DoHighlight
      \end{tikzpicture}%
    }%
    \def\SOUL@everyhyphen{%
      \discretionary{%
        \SOUL@setkern\SOUL@hyphkern
        \SOUL@sethyphenchar
        \tikz[overlay, remember picture] \highlight@EndHighlight ;%
      }{%
      }{%
        \SOUL@setkern\SOUL@charkern
      }%
    }%
    \def\SOUL@everyexhyphen##1{%
      \SOUL@setkern\SOUL@hyphkern
      \settowidth{\item@width}{##1}%
      \makebox[\item@width]{}%
      \discretionary{%
        \tikz[overlay, remember picture] \highlight@EndHighlight ;%
      }{%
      }{%
        \SOUL@setkern\SOUL@charkern
      }%
    }%
    \def\SOUL@everysyllable{%
      \begin{tikzpicture}[overlay, remember picture]
        \path let \p0 = (begin highlight), \p1 = (0,0) in \pgfextra
          \global\highlight@previous=\y0
          \global\highlight@current =\y1
        \endpgfextra (0,0) ;
        \ifdim\highlight@current < \highlight@previous
          \highlight@DoHighlight
          \highlight@BeginHighlight
        \fi
      \end{tikzpicture}%
      \settowidth{\item@width}{\the\SOUL@syllable}%
      \makebox[\item@width]{}%
      \tikz[overlay, remember picture] \highlight@EndHighlight ;%
    }%
    \SOUL@
  }
\else
  \newcommand{\anonymize}[1]{#1}
\fi

\newcommand{\linebreakand}{%
  \end{@IEEEauthorhalign}
  \hfill\mbox{}\par
  \mbox{}\hfill\begin{@IEEEauthorhalign}
}

\makeatother

\title{\LARGE \bf Real-time Robotics Situation Awareness for Accident Prevention in Industry}

\author{\IEEEauthorblockN{1\textsuperscript{st} Juan M. Deniz}
\IEEEauthorblockA{\textit{Robotics and Artificial Inteligence Lab }\\
\textit{Technological University of Uruguay}\\
Rivera, Uruguay\\
juan.deniz@utec.edu.uy}
\and
\IEEEauthorblockN{2\textsuperscript{nd} Andre S. Kelboucas}
\IEEEauthorblockA{\textit{Robotics and Artificial Inteligence Lab}\\
\textit{Technological University of Uruguay}\\
Rivera, Uruguay \\
andre.dasilva@utec.edu.uy}
\and
\IEEEauthorblockN{3\textsuperscript{rd} Ricardo Bedin Grando}
\IEEEauthorblockA{\textit{Robotics and Artificial Inteligence Lab} \\
\textit{Technological University of Uruguay}\\
Rivera, Uruguay \\
0000-0002-2939-5304}
}

\begin{document}

\maketitle
\thispagestyle{empty}
\pagestyle{empty}

\begin{abstract}

This study explores human-robot interaction (HRI) based on a mobile robot and YOLO to increase real-time situation awareness and prevent accidents in the workplace. Using object segmentation, we propose an approach that is capable of analyzing these situations in real-time and providing useful information to avoid critical working situations. In the industry, ensuring the safety of workers is paramount, and solutions based on robots and AI can provide a safer environment. For that, we proposed a methodology evaluated with two different YOLO versions (YOLOv8 and YOLOv5) alongside a LoCoBot robot for supervision and to perform the interaction with a user. We show that our proposed approach is capable of navigating a test scenario and issuing alerts via Text-to-Speech when dangerous situations are faced, such as when hardhats and safety vests are not detected. Based on the results gathered, we can conclude that our system is capable of detecting and informing risk situations such as helmet/no helmet and safety vest/no safety vest situations.

\end{abstract}
\vspace{5mm}


\vspace{-3mm}
% \section*{Supplementary Material}\label{supplementary_material}

% Video of the experiments available at: \anonymize{ AAAAAAAAAAAAAAAAAAAAAAAAAAAAA}
% % \texttt{\url{https://youtu.be/8az3agIXogQ}}. 
% Released code at: \anonymize{
% % \texttt{\url{https://github.com/ricardoGrando/DoCRL}}. 
% AAAAAAAAAAAAAAA}

\vspace{-2.5mm}
\section{Introduction}
\label{introduction}

In industrial environments, the workplace can face dangerous and chaotic situations for workers, resulting in frequent accidents between workers and equipment or machinery \cite{kim2021smart}. It can exist situations of an impact from an object, collision with working tools and obstacles, contact with toxic materials, and even bad positioning and posture. Many of these situations may occur based on the lack of respect for the norms of protection and/or inefficient supply of maintenance for equipment that has the potential to hurt a person physically. Therefore, being able to detect critical working situations and be aware of possible hazards can be an interesting approach to handling this situation \cite{endsley2008situation}.

Situation Awareness (SA) has been researched to detect these moments of danger and somehow provide a positive output in terms of avoiding or reducing the risks of accidents \cite{endsley2008situation}. It relies on the ability to perceive elements and events to understand better the actual situation of a scenario and the people in it, making decisions or proving it so danger and costs can be minimized. Usually attached to Human-Robot Interaction (HRI) concepts, SA allows us to visualize, identify, and alert unknown dangerous situations \cite{riley2016situation}.  

SA and HRI have already been used to tackle this problem. However, real-time solutions using mobile robots are still limited \cite{jentsch2016human, kumar2020survey, jones2020distributed, opiyo2021review}. In this paper, we propose a methodology based on HRI combined with YOLO \cite{redmon2016you} and a mobile robot for real-time detection of workplace risky situations, aware of the environment with possible problems to be faced. Figure \ref{fig:system} shows our proposed framework based on the Locobot mobile robot. Overall, the paper has the following contributions:

\begin{figure}
    \centering
    \includegraphics[width=\linewidth]{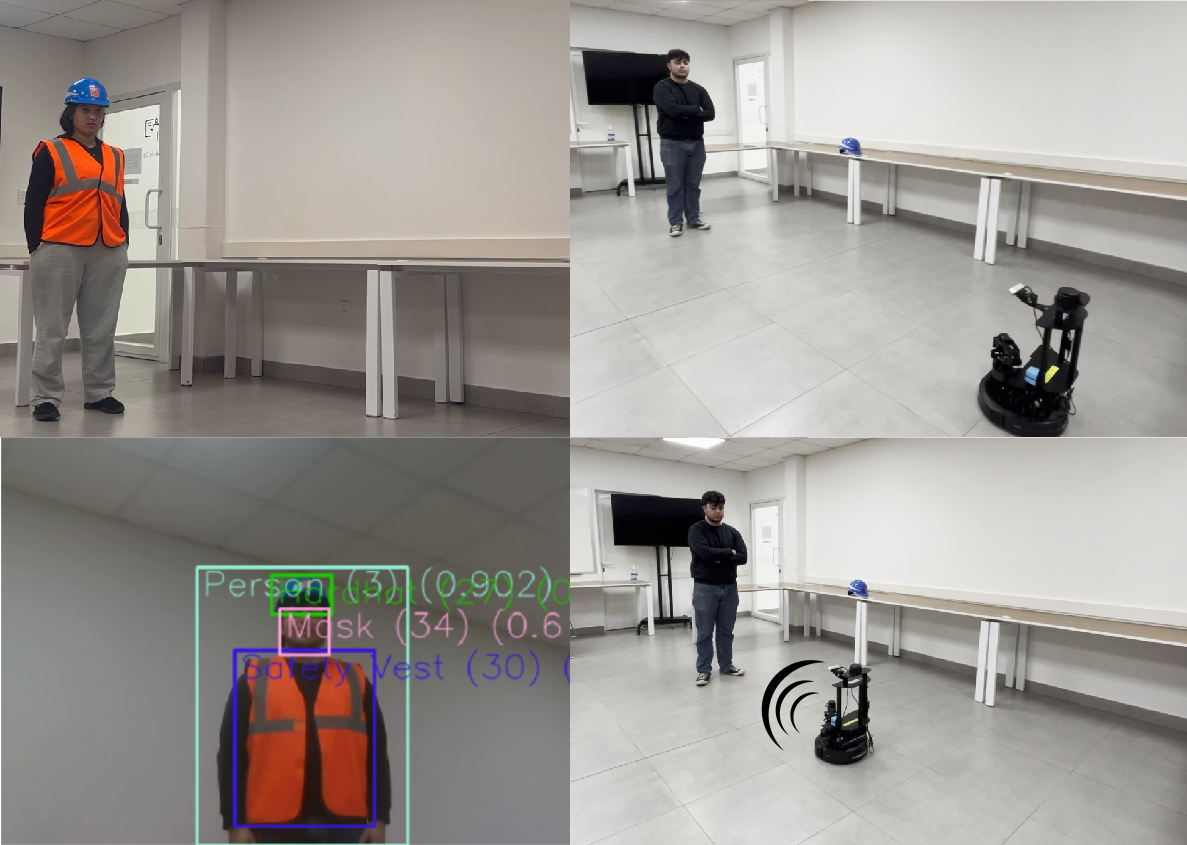}
    \caption{Our proposed system in a helmet and safety vest detection evaluation.}
    \label{fig:system}
\end{figure}

\begin{itemize}

\item We present a real-time methodology based on YOLO and a mobile robot capable of detecting risky situations that can be used for SA in industry and other environments.

\item We show that our methodology can be validated using different architectures of the YOLO framework.

\item We prove that our SA framework works in a vest/no vest situation and that it can be scaled to more situations.

\end{itemize}

This paper is organized as follows: the related works section (Sec. \ref{related_works}) is presented in the sequence. We show our methodology in Sec. \ref{methodology} and the results are presented in Sec. \ref{results}. Lastly, we highlight our contributions and present future works in Sec. \ref{conclusion}.

\section{Related Work}
\label{related_works}

A couple of works have already addressed the topic of situation awareness for industrial demands \cite{salmon2009measuring, nazir2012role}, robotic surgery \cite{ginesi2020autonomous}, accidents analysis \cite{naderpour2015role}, military \cite{jentsch2016human} and more. Salmon et al. \cite{salmon2009measuring} performed a review on SA and its importance for the industry in general. It states that SA is the awareness that a person has of a situation, such as an operator, to understand what is going on in a specific scenario in time. It discusses many methods and applies some of them in examples with people in specific scenarios to validate it. This work addresses this topic at a nontechnological level, addressing SA as perceived and analyzed by humans. 

Nazir et al. \cite{nazir2012role} also address SA as its fundamental idea derived from armies in the First World War. It describes it by proposing SA at three levels: Perception, Comprehension, and Projection. In the industrial context, it argues that a human operator must have shared knowledge, collective dynamic understanding, and shared mental modeling. It concludes that SA plays a determinant role in the industrial process in reducing abnormal situations that can lead to accidents.

From this conceptual interpretation of SA, many works have used SA in Robotics, Artificial Intelligence and Computer Vision concepts, looking for the development of intelligent systems that could work as an operator or something similar in industry and other scenarios. Riley et al. \cite{riley2004hunt} used SA concepts to apply them to vehicle control in robotics and Human-Robot Interaction (HRI), specifically for tasks related to search and rescue. They seek to look for the information needed to be used during a victim search. It concluded by mentioning that it had difficulties developing it, given the lack of support and data acquisition problems.  

Naderpour et al. \cite{naderpour2015role} proposed the application of the SA Error Taxonomy methodology to analyze the effect in three different contexts of accidents. The analysis performed showed that the accidents were due to a lack of appropriate design of operator support systems, errors due to poor mental models, and that there is a need to develop technological systems to lower operator workload and stress and avoid human errors.

Jones et al. \cite{jones2020distributed} \cite{jentsch2016human} proposed a methodology of Distributed Situational Awareness to be used in multi-robot systems. The proposed methodology aimed to rapidly and accurately capture an environment and, based on it, act in order to minimize a problem. However, it concluded that this conceptual idea requires advanced hardware and algorithms that demonstrate that it is safe and reliable and perform testing with a swarm of robots.

SA has already been used in surgical tasks \cite{ginesi2020autonomous, onyeogulu2022situation} Ginesi et al. \cite{ginesi2020autonomous} presented an idea of SA as a framework to implement surgical task automation, where multiple actions need to be performed, and the sequence of execution is not predefined. The proposed SA module was used to detect the works as a semantic interpretation of the sensing information. It provides middleware with a high-level description of the scenario in real-time. Onyeogulu et al. \cite{onyeogulu2022situation} used the YOLO to perform automated invasive surgery. It analyzed four different versions, concluding that the v5 version provided the best results.

Based on these works, we present a methodology focused on real-time situation awareness in a simulated industrial environment for accident prevention. We show that our approach is based on acquiring information from a YOLO model and converting it to commands to a mobile robot; it is possible to detect and inform about risky situations using the robot as a robotic operator.

\section{Methodology}
\label{methodology}

In our methodology, we introduce our proposed SA system for real-time accident detection. Firstly, we introduce our robotic platform and the YOLO models used to perform object detection alongside the dataset used to train the models. Then, we present our robotic solution based on the Locobot robot and a ROS2 package that performs real-world detection and HRI with the user. 

\subsection{LoCoBot and ROS2}

LoCoBot is a mobile robot platform that provides a benchmark for robotics applications. Our version of the robot has a mobile base that is capable of navigating and performing localization. The robot also has a 360-degree Lidar and a depth camera that is used for localization, navigation, and more. The robot has an onboard computer that allows it to perform onboard processing, such as object detection, using YOLO. For this research, we also added a speaker so the robot could provide commands and information in the form of audio. 

\begin{figure}[!h]
    \centering
    \includegraphics[width=0.5\linewidth]{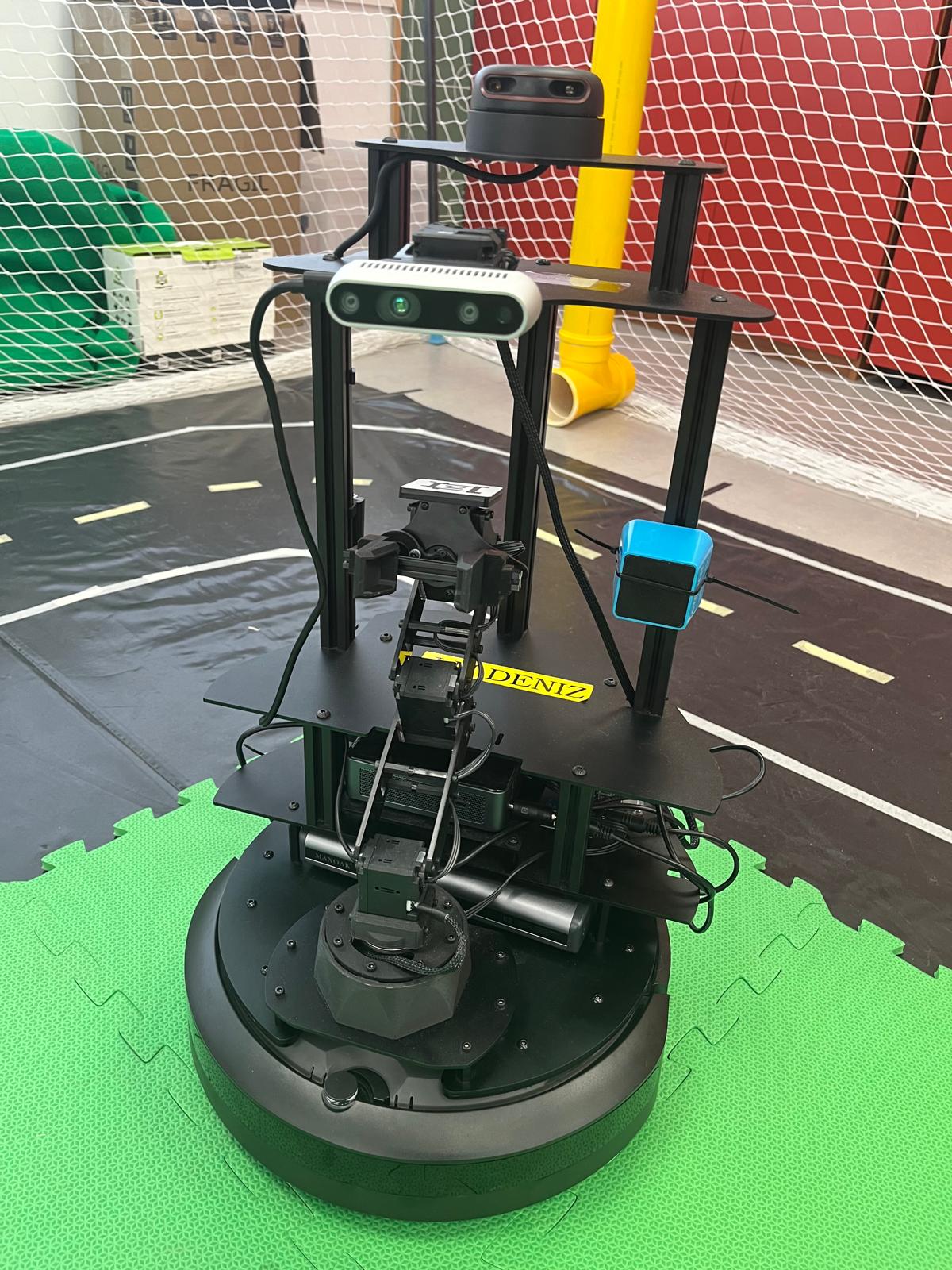}
    \caption{LoCoBot - Robot used for this research}
    \label{fig:locobot}
\end{figure}

The robot's system is based on ROS2 framework \cite{doi:10.1126/scirobotics.abm6074}. It provides a set of tools to develop robotics applications. It is open source, providing state-of-the-art support for users to develop their applications. It was improved from ROS1, providing a system that does not rely on a centralized master node. For this work, all applications were developed based on the Locobot platform and ROS2 framework. All applications were created to run locally on the robot without using external processing. Our robot used for the evaluation of this work can be seen in Figure \ref{fig:locobot}.

\subsection{YOLO}

To perform the object detection, we used two variations of the YOLO framework \cite{redmon2016you}, the version YOLOv8 and YOLOv5. It was selected based on the use of these versions in previous papers related to SA \cite{onyeogulu2022situation}. 

YOLOv5 employs a single deep neural network to predict objects in the form of bounding boxes within an image. It was developed by Ultralytics, following the past version of the framework. The framework was created to provide real-time object detection, with pre-trained models that have, in most cases, dozens of categories of objects that can detect. All versions, including the v5 version, are trained on a large dataset that contains millions of images. To achieve that, Yolo relies on a model that usually has multiple convolutions that extract features from the images, adding bonding boxes of the probability of a specific object within the image. 

Overall, both YOLOv5 and YOLOv8 versions have the FNP backbone, neck, and head pieces. YOLOv5 and YOLOv8 differ greatly in the head module. The coupling structure was proposed in YOLOv5 version, which was changed to a decoupling one in YOLOv8. Also, YOLOv8 has an anchor-free model, while the YOLOv5 is an anchor-based model. This feature impacts the complexity of the model, where the Anchor-based model helps the model to detect better objects of different sizes and aspect ratios. For YOLOv8,  by eliminating the need to define anchor boxes, it gets a less complex model. However, besides that, both versions have similar performance.

\subsection{Safety Equipment Dataset}

For this application, we used a multiclass dataset for classification and object detection, called "Construction Site Safety Dataset" \cite {construction-site-safety_dataset}. It contains 2,605 training images, 114 validation images, and 82 test images with annotations in YOLOv8 format. The dataset includes 10 classes as follows: Hardhat, Mask, NO-Hardhat, NO-Mask, NO-Safety Vest, Person, Safety Cone, Safety Vest, Machinery, and Vehicle. This dataset provides the advantage of having several important classes for applications related to this work. 

Additionally, the dataset employs some augmentation techniques that enhance the training process, including randomly inserting black pixels, combining four images, stretching, and applying zoom effects. These augmentation techniques can help to create a diverse set of training examples, improving the model's ability to generalize to new unseen data. This aspect of the dataset is particularly advantageous for SA, as it allows the trained model to learn from a wide array of scenarios, such as different environmental conditions, various object interactions, and unexpected obstacles.

\subsection{Real Time Implementation}

\begin{figure}[!h]
    \centering
    \includegraphics[width=\linewidth]{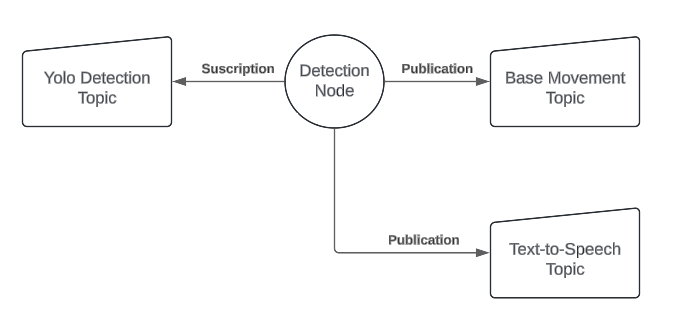}
    \caption{Detection node scheme}
    \label{fig:node-detection}
\end{figure}

For the system created, we used ROS2 Humble \cite{doi:10.1126/scirobotics.abm6074} for the implementation. The workspace includes packages to operate the Realsense camera, the YOLOv8 package with a detection node utilizing this camera's topic, a package for Locobot base movement, and a package enabling text-to-speech functionality. Additionally, a package has been developed that includes a node utilizing the packages mentioned above, as shown in Figure \ref{fig:node-detection}.

\begin{figure}[!h]
    \centering
    \subfloat[Experimental scenario.]{\includegraphics[width=0.46\linewidth]{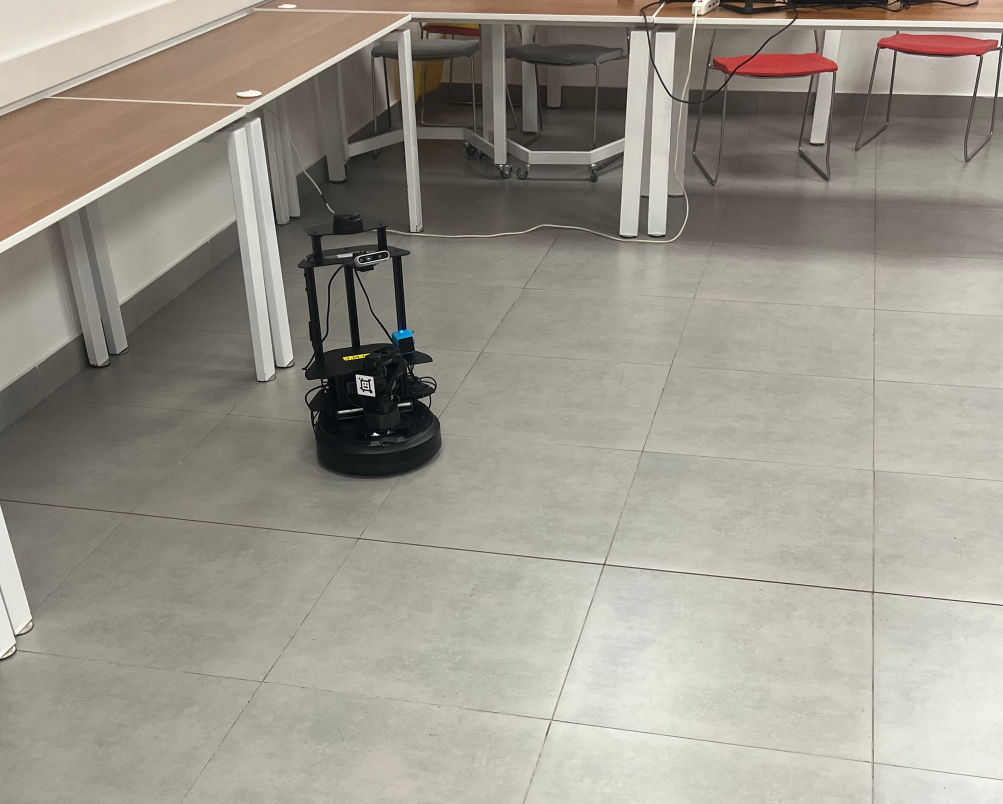}\label{fig:img_scenario}}
    \hspace{1mm} % Adds a horizontal space of 1cm between the images.
    \subfloat[Experimental scenario and test subject.]{\includegraphics[width=0.48\linewidth]{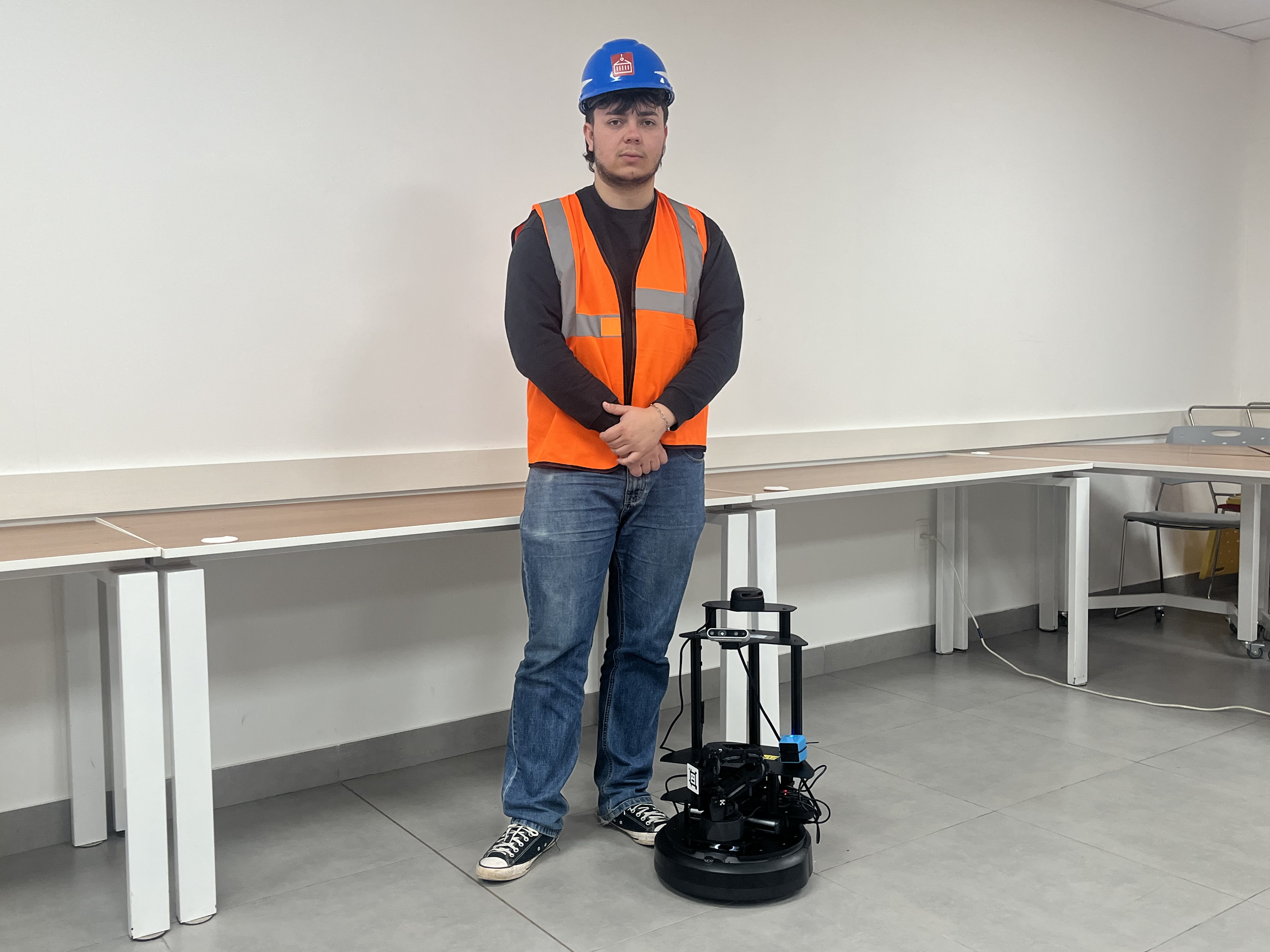}\label{fig:img_scenario2}}
    \caption{Scenario used for the tests, showing the test subject with the safety equipment.}
    \label{fig:scenario}
    \vspace{-5mm}
\end{figure}

This node subscribes to the YOLO detection topic to make decisions and publish to the movement and Text-to-Speech topics. This node includes the routine to move the Locobot when a risk situation is detected; for example, in the absence of safety equipment, the robot stops and emits an alert.

\subsection{Experimental Setup}
\label{experimentalsetup}
For the experiments, a 2x3 meter area was used with two different test subjects, each wearing safety equipment, as shown in Figure \ref{fig:scenario}. The experiments included five types of trials:
\begin{itemize}
    \item Experiment 1: Static subject wearing safety equipment.
    \item Experiment 2: Static subject without safety equipment.
    \item Experiment 3: Moving subject wearing safety equipment.
    \item Experiment 4: Moving subject without safety equipment.
    \item Experiment 5: Both test subjects, one wearing safety equipment and the other without.
\end{itemize}

\section{Experimental Results}
\label{results} % Reescrito por RIcardo

In this section, the results and metrics obtained for the two trained models, YOLOv5s and YOLOv8m, are presented.

\begin{figure*}[!h]
    \centering
    \subfloat[F1-Curve YOLOV8]{\includegraphics[width=0.2\linewidth]{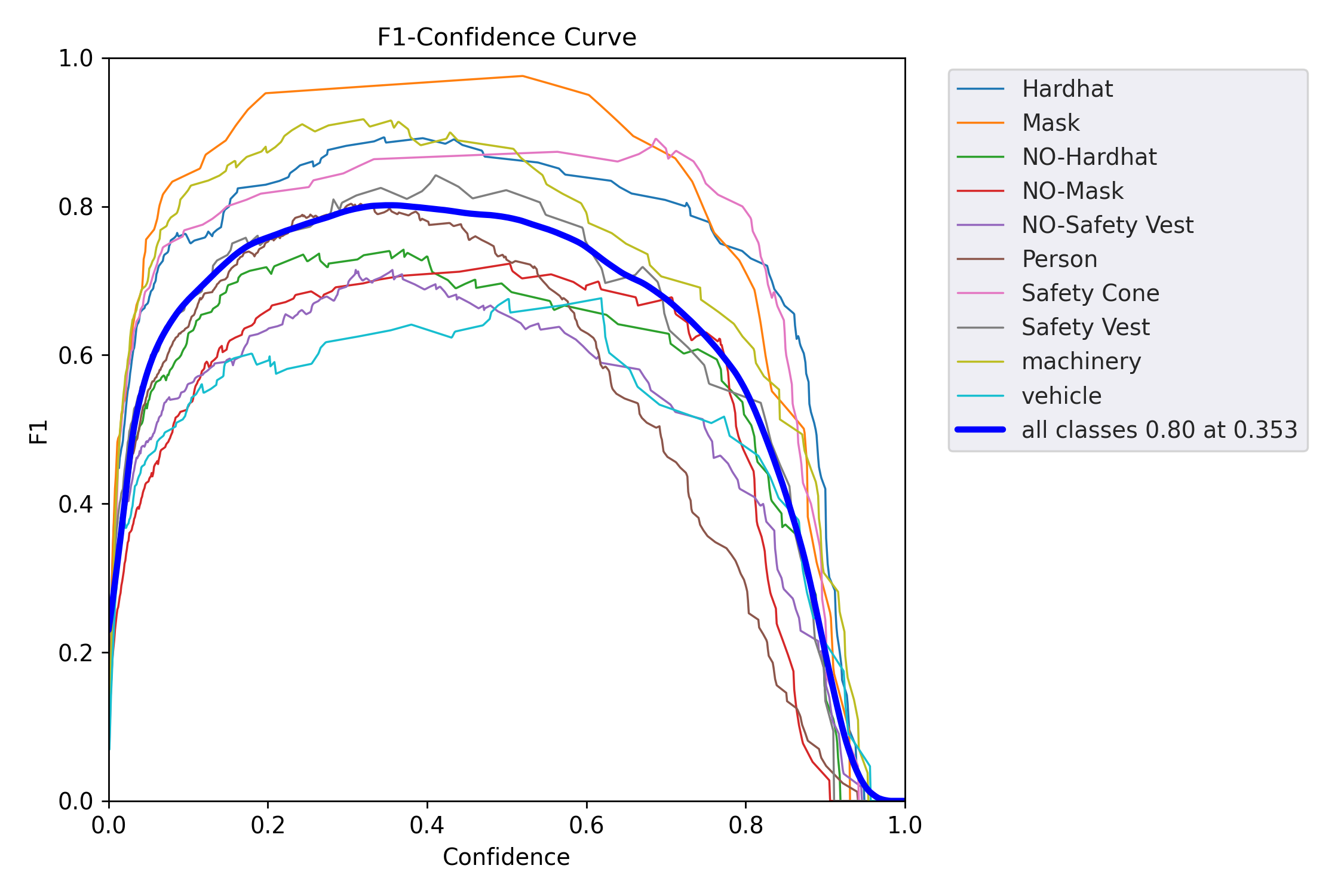}\label{fig:F1CurveV8}}
    \subfloat[Precision-Confidence Curve YOLOV8]{\includegraphics[width=0.2\linewidth]{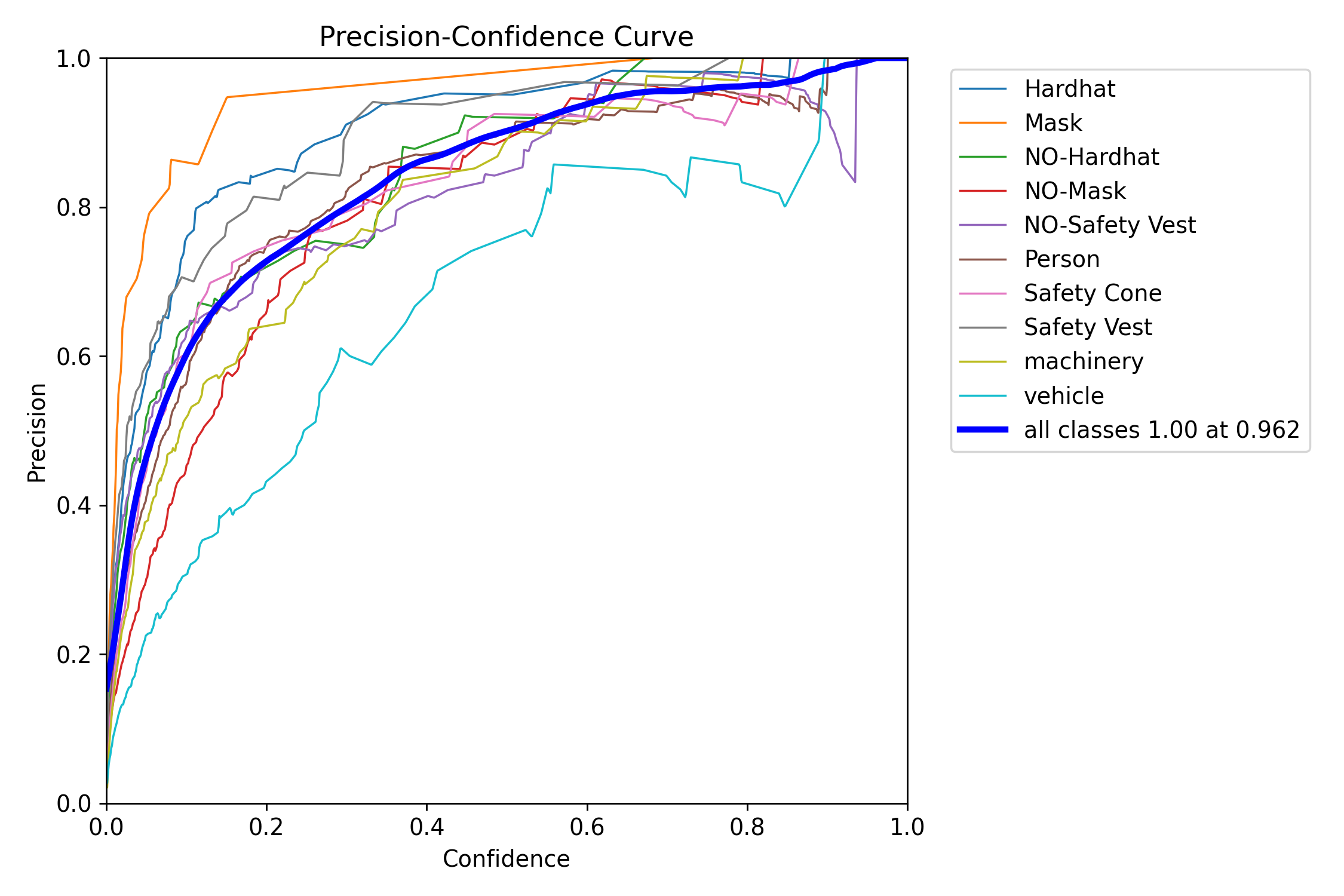}\label{fig:PCurveYOLOV8}}
    \subfloat[Precision-Recall Curve YOLOV8]{\includegraphics[width=0.2\linewidth]{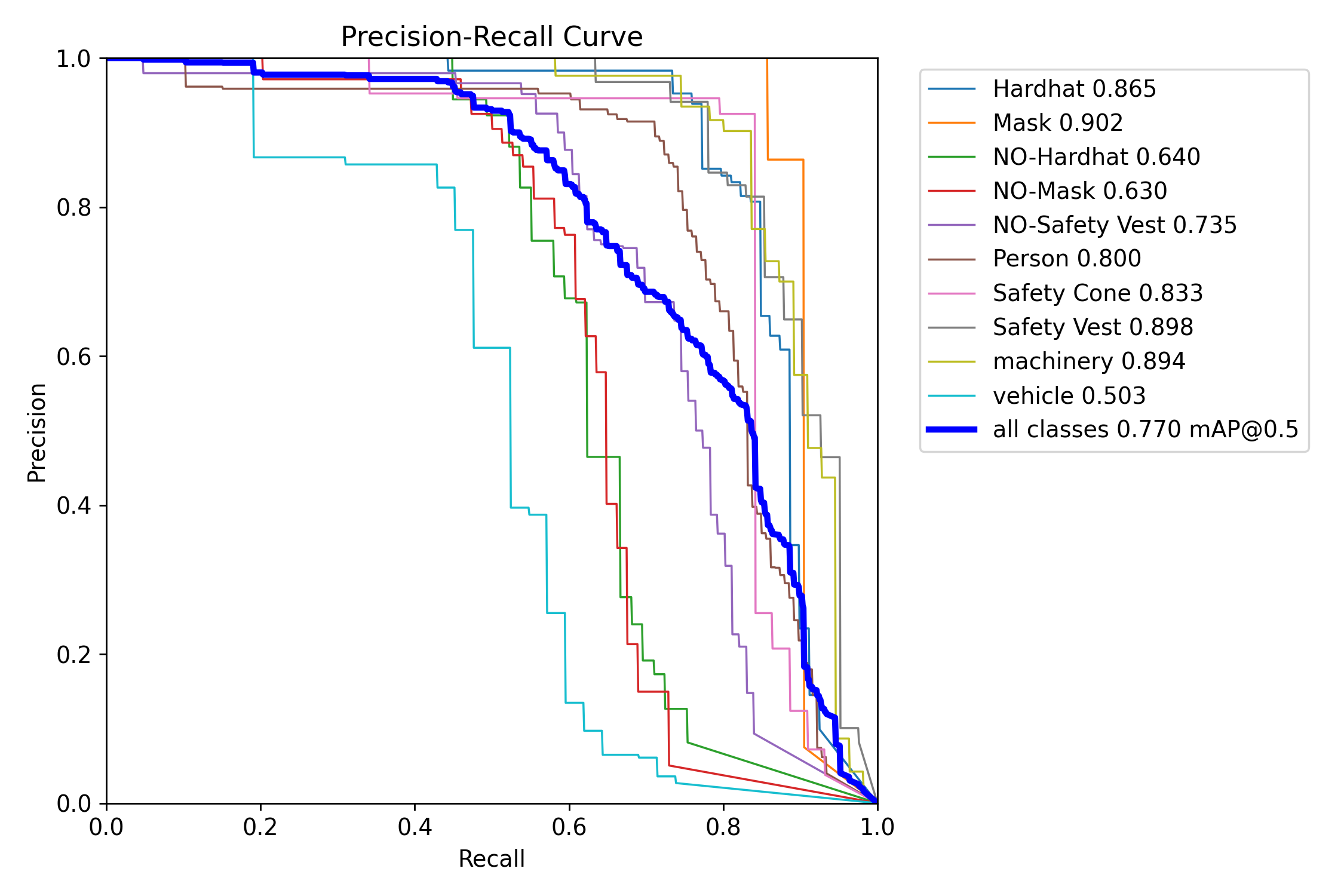}\label{fig:PRYOLOV8}}
    \subfloat[Recall-Confidence Curve YOLOV8]{\includegraphics[width=0.2\textwidth]{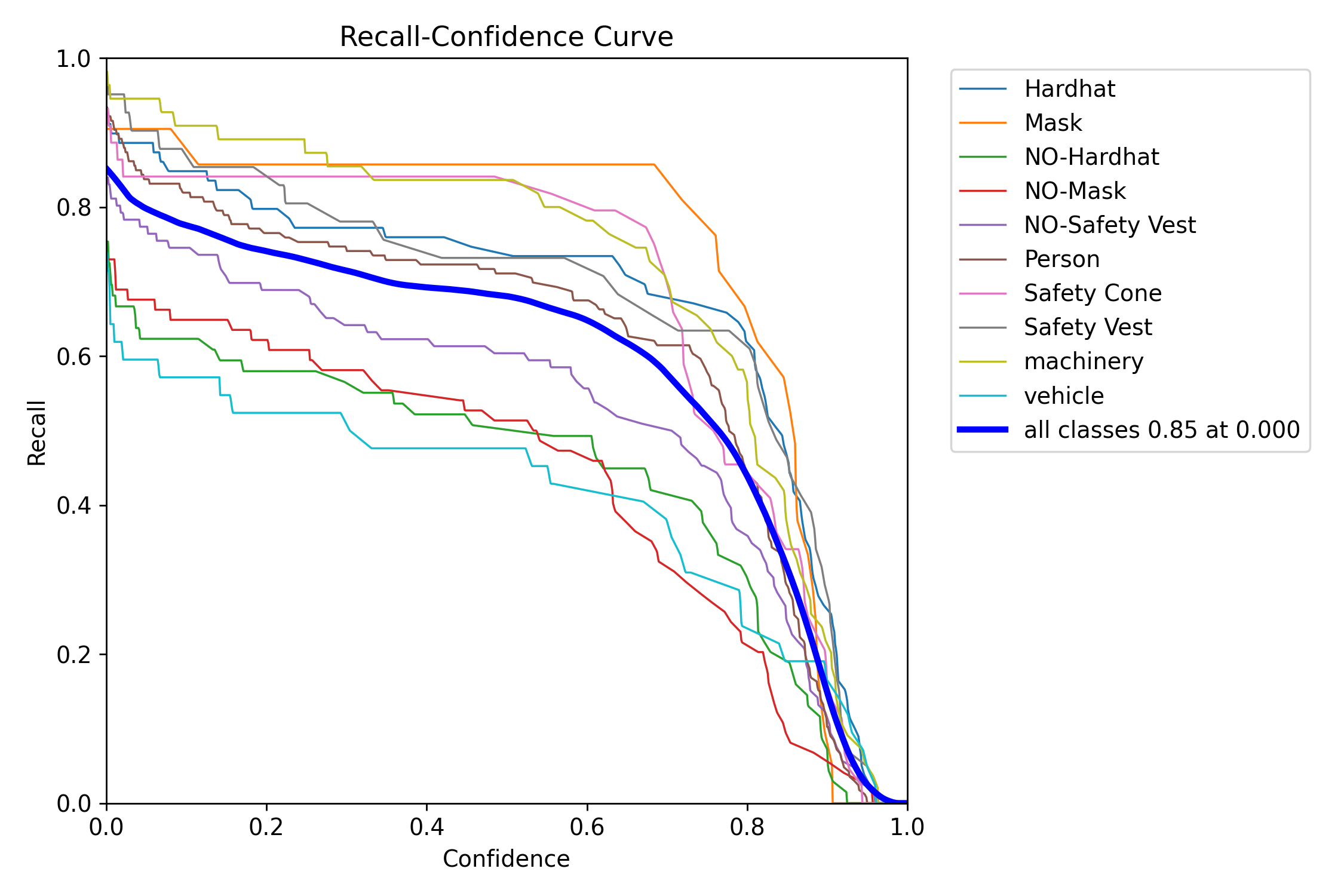}\label{fig:R_Curve YOLOV8}} \\
    \subfloat[F1-Confidence Curve YOLOV5]{\includegraphics[width=0.2\textwidth]{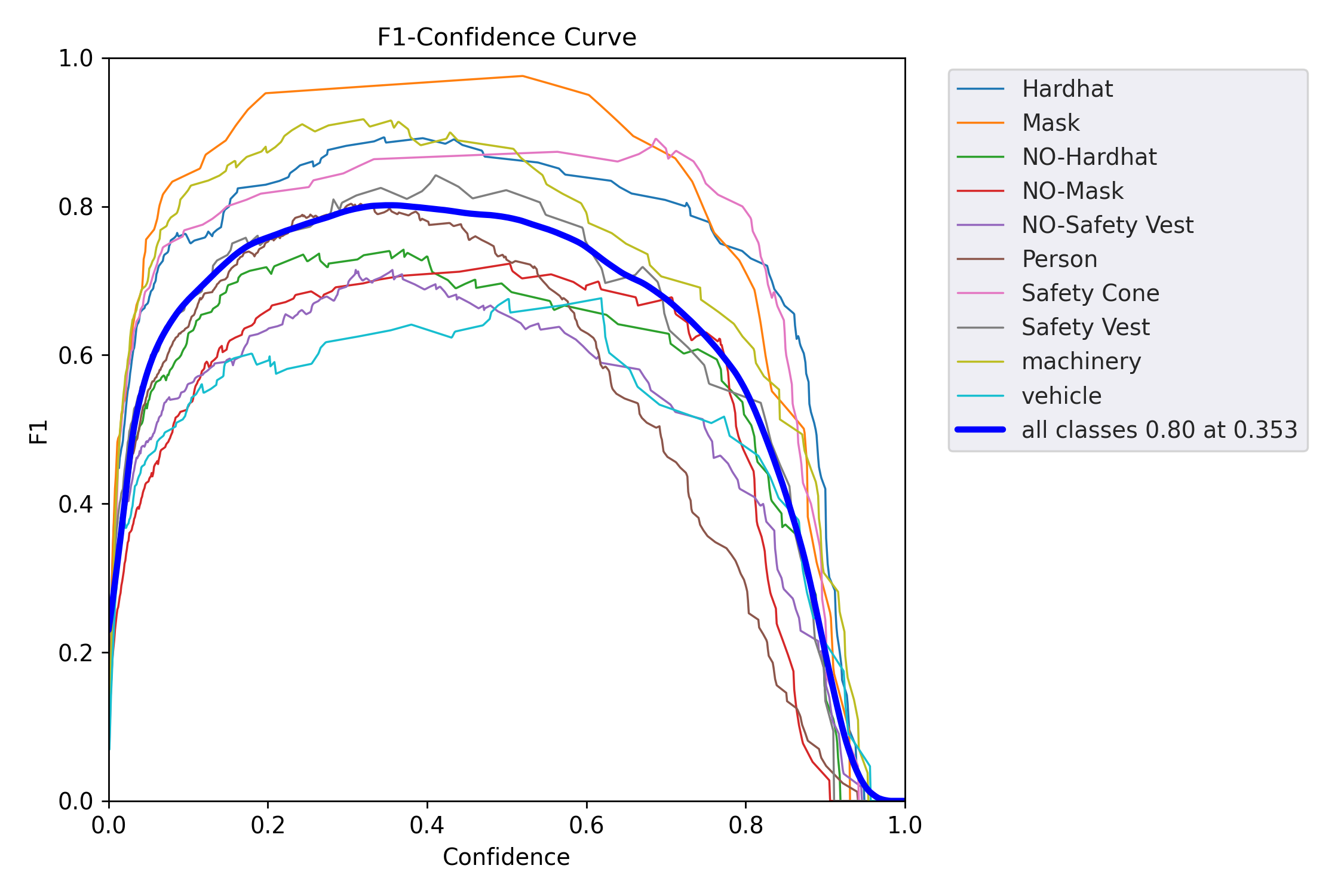}\label{fig:F1CurveV5}} 
    \subfloat[Precision-Confidence Curve YOLOV5]{\includegraphics[width=0.2\textwidth]{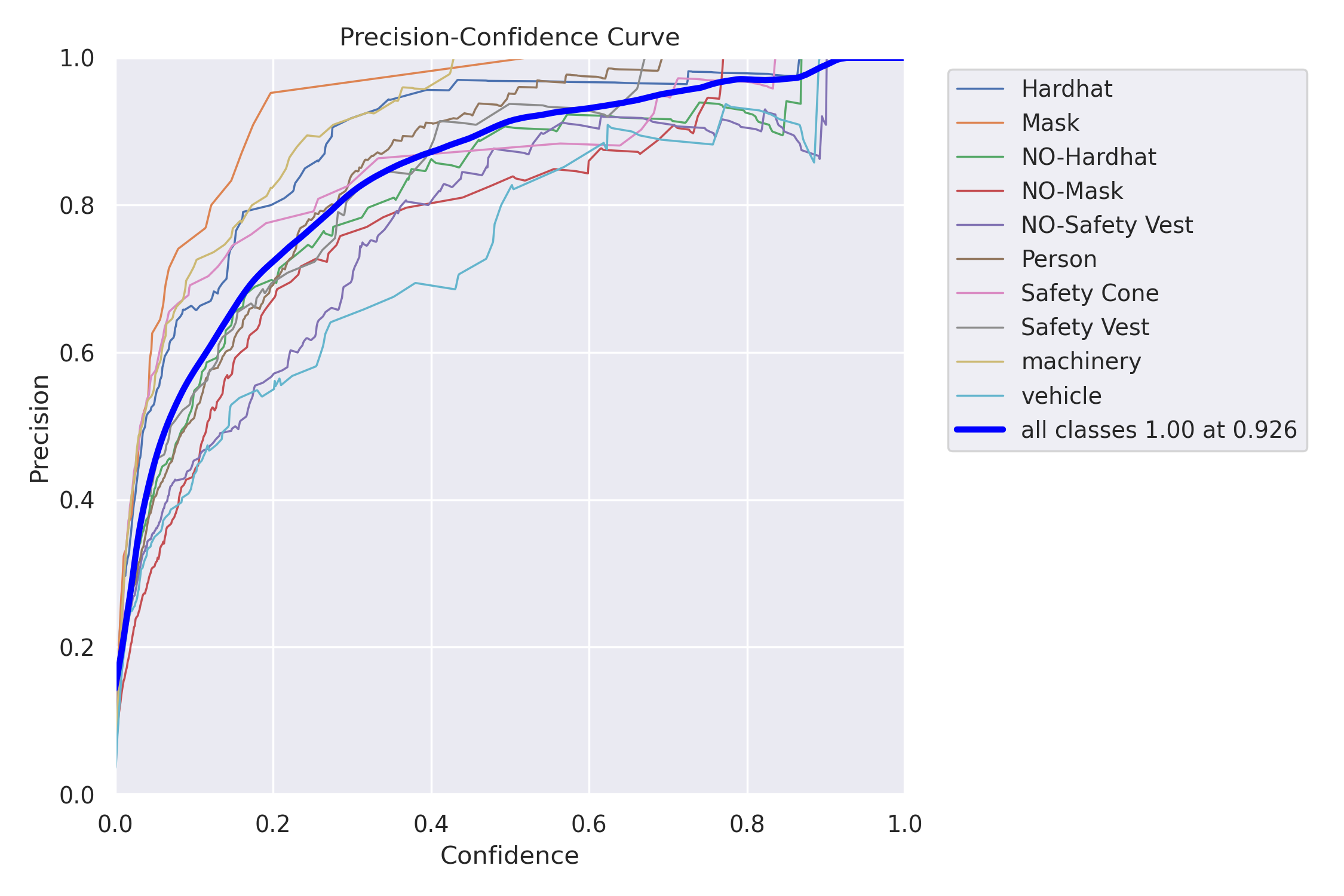}\label{fig:PCurveV5}}
    \subfloat[Precision-Recall Curve YOLOV5]{\includegraphics[width=0.2\textwidth]{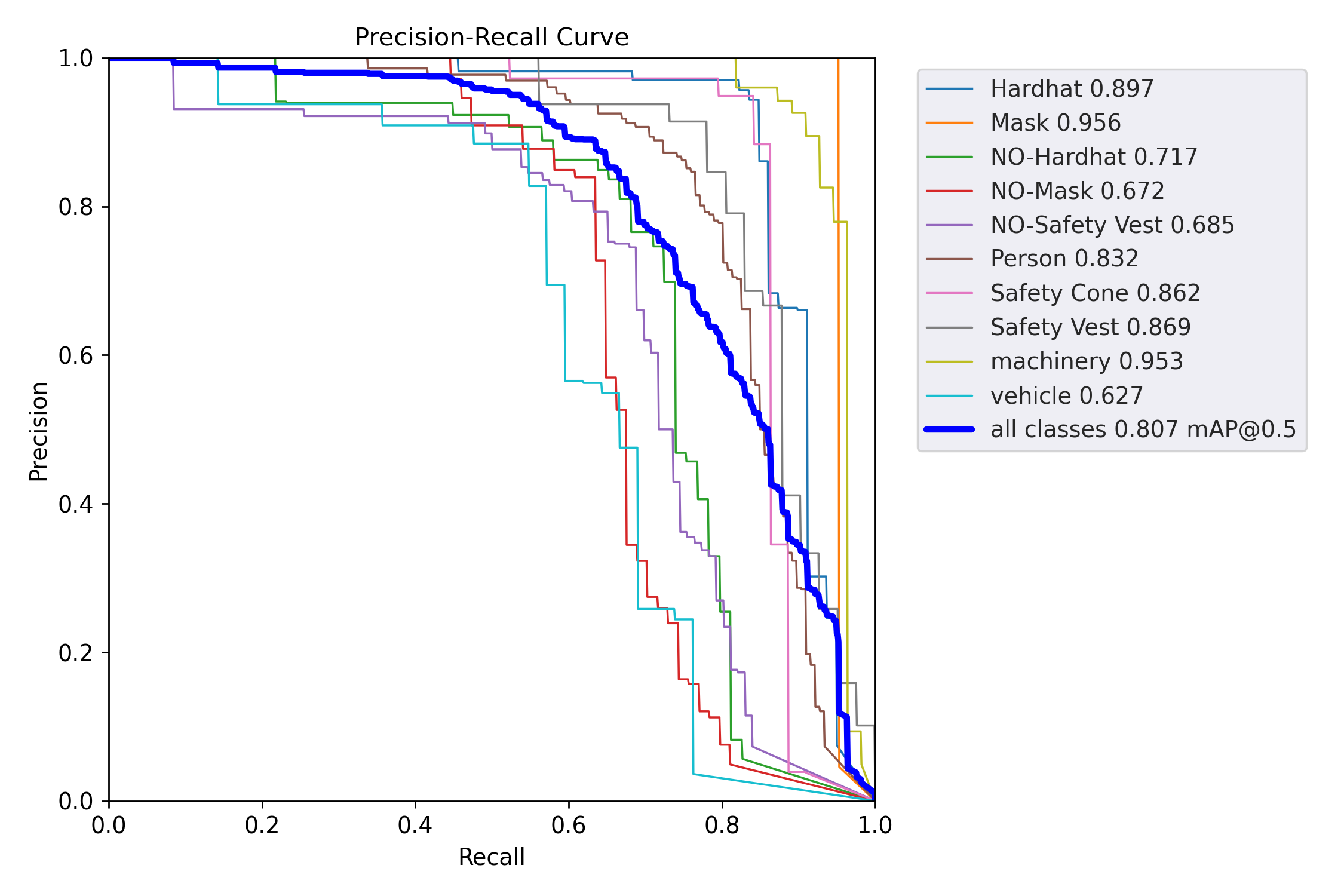}\label{fig:PRCurveYOLOV5}}
    \subfloat[Recall-Confidence Curve YOLOV5]{\includegraphics[width=0.2\textwidth]{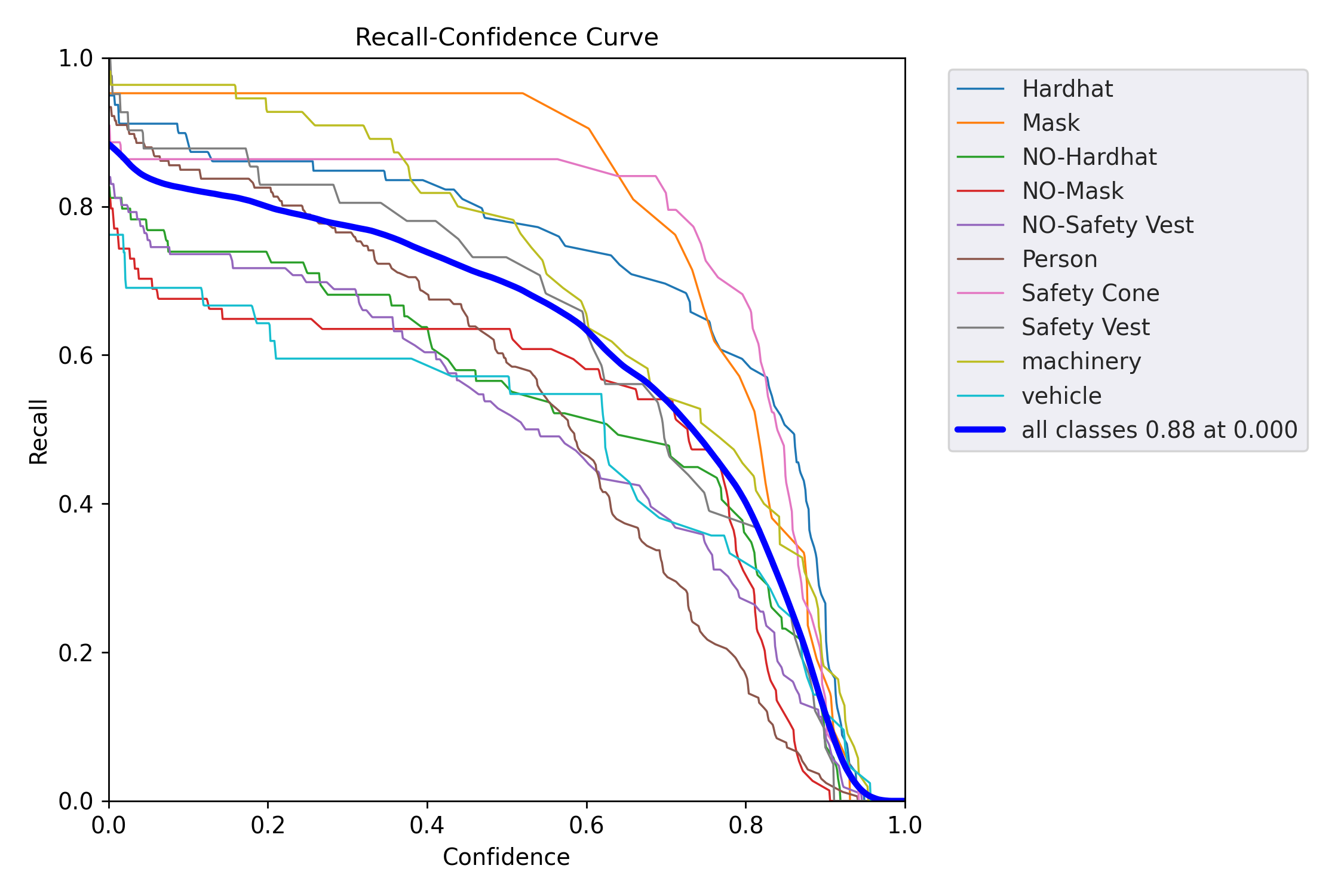}\label{fig:RCuveV5}}
    \caption{F1-Curve, Precision-Confidence, Precision-Recall and Recall-Confidence Curves}
    \label{fig:environment1}
    \vspace{-5mm}
\end{figure*}

\begin{figure}[ht!]
    \centering
    \includegraphics[width=1\linewidth]{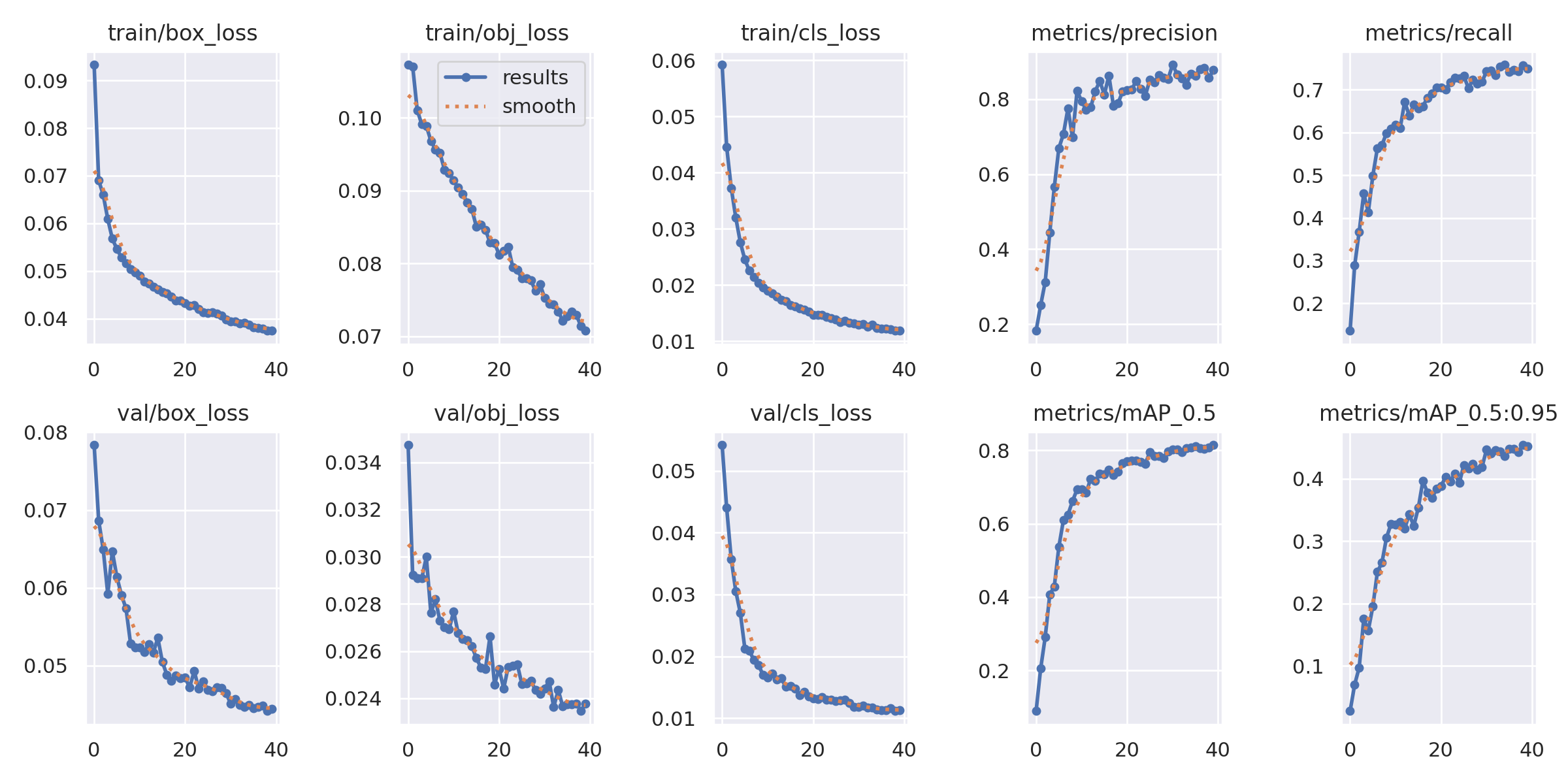}
    \caption{YoloV5 training results.}
    \label{fig:yolov5results}
\end{figure}

\begin{table}[!h]
    \centering
    \begin{tabular}{|c|c|c|c|} \hline 
 Experiment& Number Tests& Tests Subject&Success rate\\ \hline 
         1&  6&  1& 100\%\\ \hline 
         2&  6&  1& 100\%\\ \hline 
         3&  4&  1& 100\%\\ \hline 
         4&  4&  1& 75\%\\ \hline 
         5&  5&  1 and 2& 80\%\\ \hline 
         1&  6&  2& 100\%\\ \hline 
         2&  6&  2& 50\%\\ \hline 
         3&  4&  2& 75\%\\ \hline 
         4&  4&  2& 25\%\\ \hline
    \end{tabular}
    \caption{Results of experiments, number of tests and success rate with all the trials}
    \label{tab:experimentalsetuptab}
\end{table}

\subsection{Results} % Reescrito por Ricardo

\begin{figure}[!h]
    \centering
    \subfloat[Subject without safety equipment.]{\includegraphics[width=0.46\linewidth]{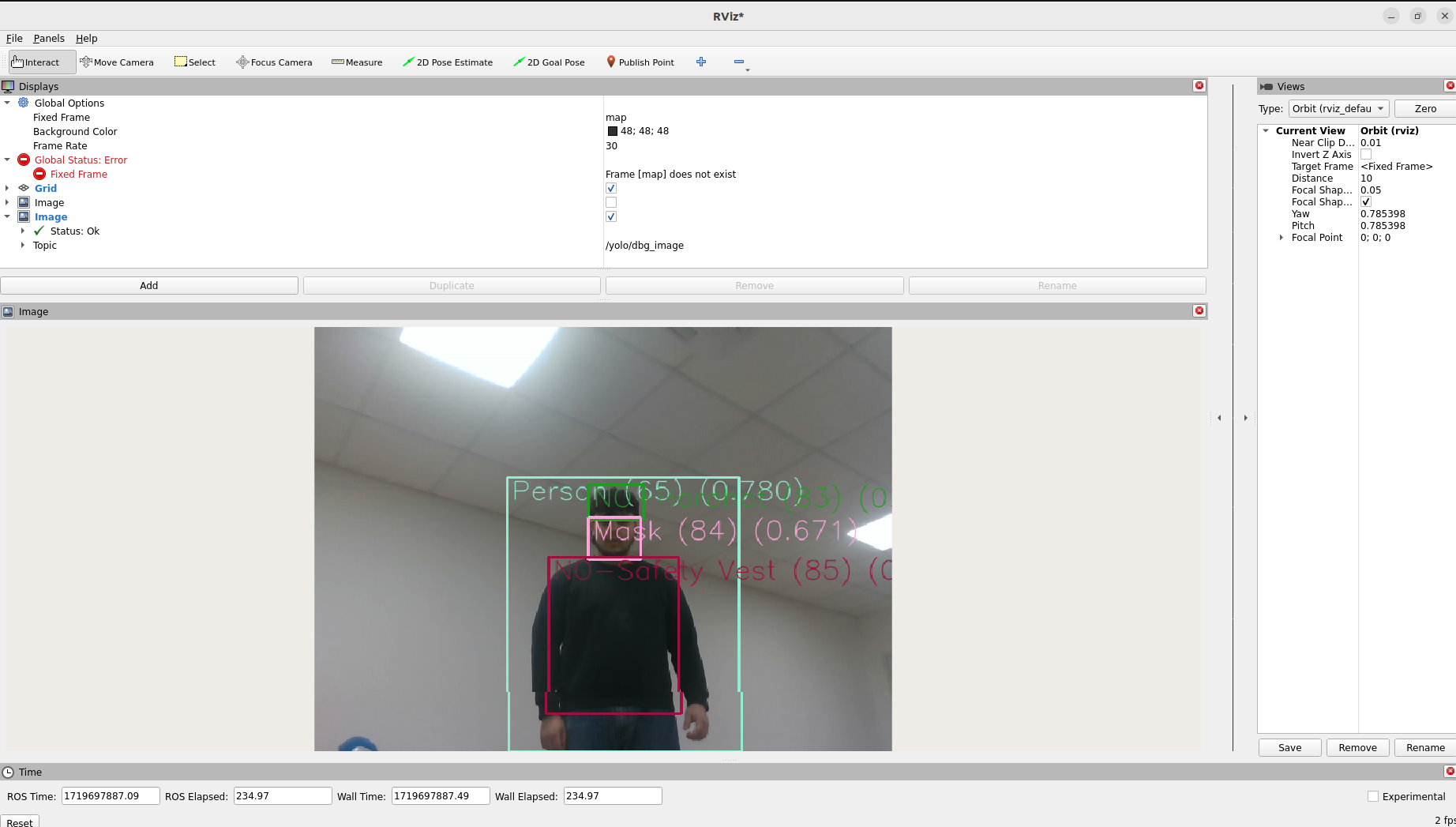}\label{fig:no safety equipment}}
    \hspace{1mm} % Adds a horizontal space of 1cm between the images.
    \subfloat[Subject wearing safety equipment.]{\includegraphics[width=0.48\linewidth]{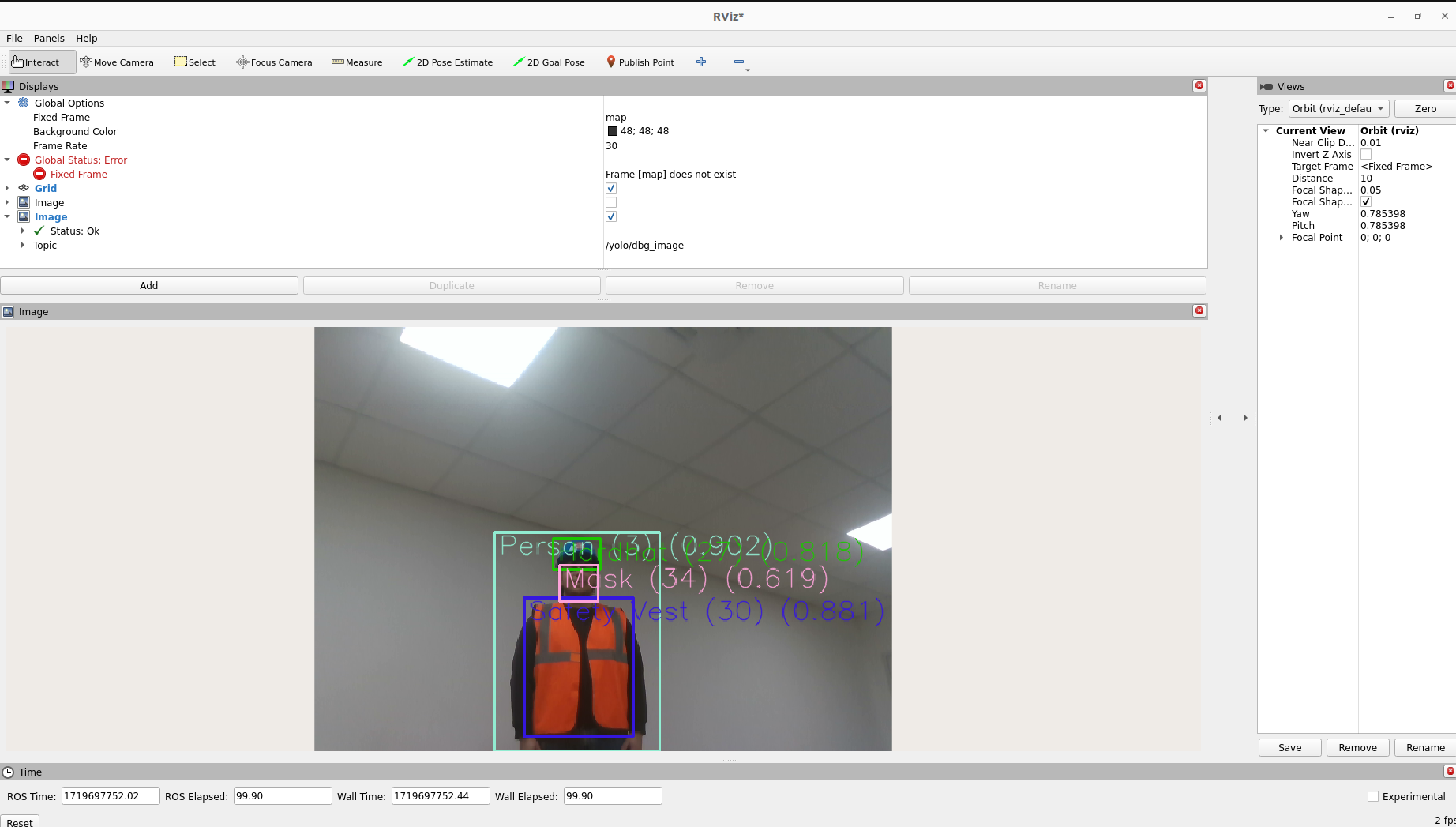}\label{fig:safety equipment}}
    \caption{Screenshots of RViz2 showing the detection topic of two different situations.}
    \label{fig:detection}
    \vspace{-5mm}
\end{figure}

\begin{table}[b!] % Use table for a single column
    \centering
    \setlength{\tabcolsep}{4pt} % Reducing the space between columns
    \caption{Box mAP50 for all models and classes.}
    \begin{tabular}{lp{1.5cm}p{1.5cm}p{1.5cm}p{1.5cm}}
        \toprule
        \textbf{Architecture} & \textbf{Hardhat}& \textbf{NO-Hardhat} &\textbf{NO-Mask} &\textbf{NO-Safety Vest} \\
        \midrule
        YOLOv8m & 0.865& 0.640& 0.630& 0.735 \\
        YOLOv5s& \textbf{0.897}& \textbf{0.717}& \textbf{0.672}& 0.685\\
        \bottomrule
    \end{tabular}\\[10pt] % Add some vertical space between tables
    \begin{tabular}{lp{1.5cm}p{1.5cm}p{1.5cm}p{1.5cm}}
        \toprule
        \textbf{Architecture} & \textbf{Person} &\textbf{Safety Cone}   &\textbf{Safety Vest} & \textbf{Mask}\\
        \midrule
        YOLOv8m & 0.800& 0.833 &0.898 & 0.902 \\
        YOLOv5s& \textbf{0.832}&  \textbf{0.862}&0.869& \textbf{0.956} \\
        \midrule
        \textbf{Architecture} & \textbf{machinery} & \textbf{vehicle} & & \\
        \midrule
        YOLOv8m & 0.894& 0.503 & & \\
        YOLOv5s& \textbf{0.953}& \textbf{0.627} & & \\
        \bottomrule
    \end{tabular}
    \label{tab:map50-values-box}
\end{table}

We can also see the resulting metrics for each trained model, such as the F1-Confidence Curve, Precision-Confidence Curve, Precision-Recall Curve, and Recall-Confidence Curve \cite{yolometrics}. 
    \begin{itemize}
        \item \textbf{F1-Confidence Curve}: Helps understand how prediction confidence affects precision and recall. Figure \ref{fig:F1CurveV8} and \ref{fig:F1CurveV5} 
        \item \textbf{Precision-Confidence Curve}: Allows evaluating the confidence of predictions made by the model. Figure \ref{fig:PCurveYOLOV8} and \ref{fig:PCurveV5}
        \item  \textbf{Precision-Recall Curve}: Shows how improving precision typically reduces recall. Figure \ref{fig:PRYOLOV8} and \ref{fig:PRCurveYOLOV5}
        \item  \textbf{Recall-Confidence Curve}: Helps understand how confidence in predictions affects the model's ability to correctly detect all positive cases, including those with low confidence. Figure \ref{fig:R_Curve YOLOV8} and \ref{fig:RCuveV5}

    \end{itemize}

\begin{figure}[ht!]
    \centering
    \includegraphics[width=1\linewidth]{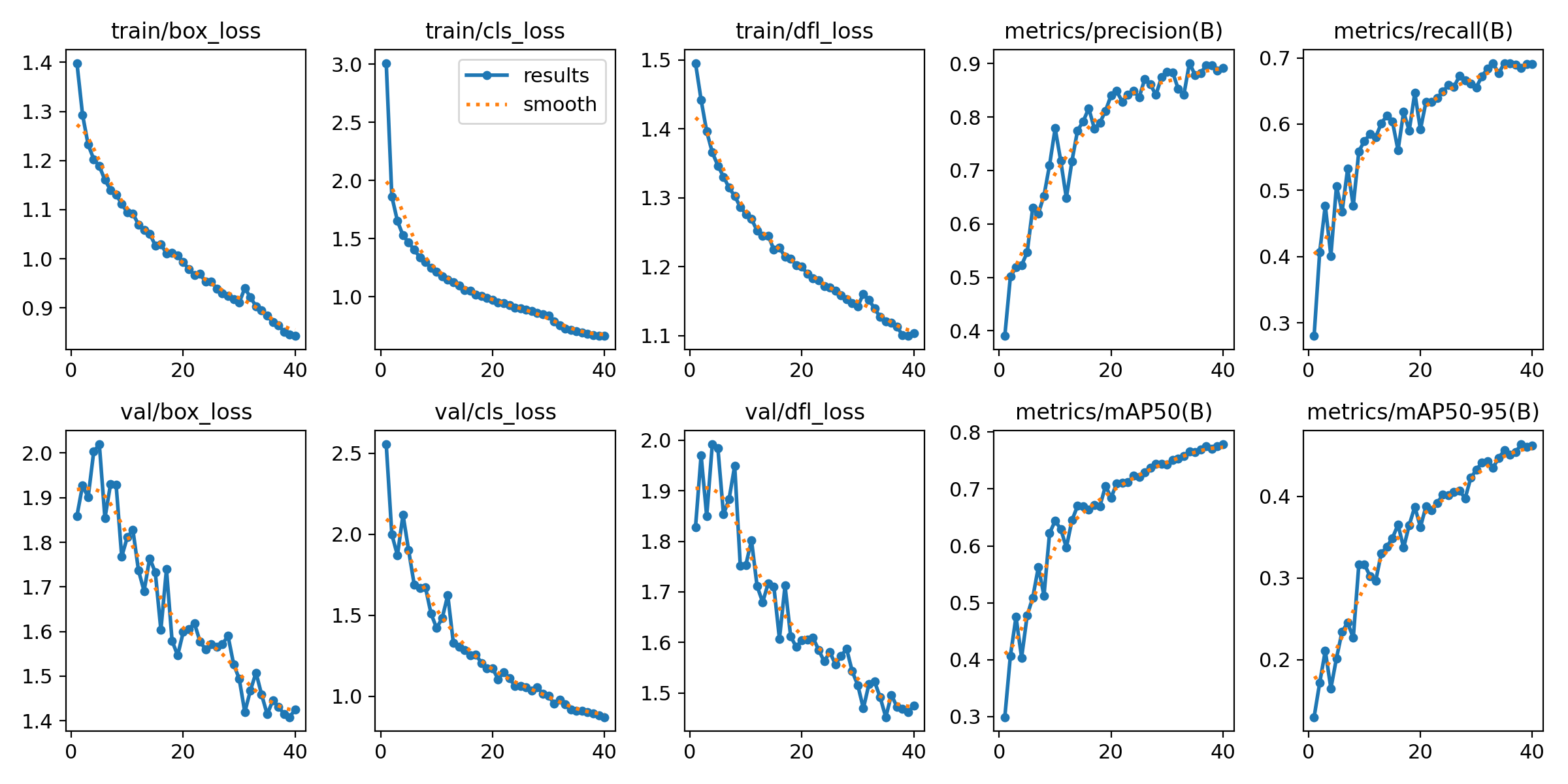}
    \caption{YoloV8 training results.}
    \label{fig:yolov8results}
\end{figure}

Additional performance graphics of both trained models can be observed in Figures \ref{fig:yolov5results} and \ref{fig:yolov8results}, showing how the loss decreases and how the precision and recall increase. Additionally, for the experiments mentioned in subsection \ref{experimentalsetup}, Table \ref{tab:experimentalsetuptab} shows the results obtained in each trial.

Finally, RViz2 was used to observe the output of the detection topic from the Locobot's camera. Two screenshots have been included for comparison, showing detections of the classes Hardhat, Mask, NO-Hardhat, NO-Mask, NO-Safety Vest, Safety Vest, and Person. In Figure \ref{fig:detection}, it is possible to see a comparison of detections with and without wearing safety equipment.

% \section{Discussion}
 
% \label{discussion}

\section{Conclusion}
\label{conclusion}

Overall, we can conclude that both YOLO were able to learn and perform safety vest and helmet detection. Our system managed to detect in all five experiments proposed, excelling in most of them, working with static and moving subjects. We evaluated using both a female and a male person, with and without safety equipment, where the results with a male (Test Subject 1) were better. That can be due to the increased difficulty of the YOLO model in detecting helmets with our Test Subject 2, who had long hair. Based on the real-world experiments conducted with the YOLOv8 version, we can analyze that the proposed robotics system was able to work as an autonomous operator and perform the interaction with the user in the simulated workplace. 

It is important to highlight that the YOLOv8 model was selected for the system given its compatibility with ROS2 Humble and the Locobot system, besides the fact that the YOLOv5 version presented the best results. For future work, we aim to improve our system with more advanced object detection frameworks and also add more risky situations to increase the robots' safety awareness capabilities.
%\vspace{-4mm}
\section*{Acknowledgment}

% \anonymize{xx}

%This work was partly founded by the Coordenação de Aperfeiçoamento de Pessoal de Nível Superior (CAPES) and Conselho Nacional de Desenvolvimento Científico e Tecnológico (CNPq). 
The authors would like to thank the Technological University of Uruguay, especially the student Any Lucía Gómez da Rosa. This work was partly supported by the Becas de Investigación UTEC - Modalidad Iniciación a la investigación iniciative.

%\vspace{-2mm}
\bibliographystyle{./bibliography/IEEEtran}
\bibliography{./bibliography/IEEEabrv,./bibliography/main}

\end{document}